\documentclass[conference]{IEEEtran}
\IEEEoverridecommandlockouts
\usepackage{cite}
\usepackage{amsmath,amssymb,amsfonts}
\usepackage{algorithmic}
\usepackage{graphicx}
\usepackage{textcomp}
\usepackage{xcolor}
\def\BibTeX{{\rm B\kern-.05em{\sc i\kern-.025em b}\kern-.08em
    T\kern-.1667em\lower.7ex\hbox{E}\kern-.125emX}}

\usepackage[latin1,utf8]{inputenc}
\usepackage[hyphens]{url}
\usepackage{hyperref}
\usepackage[keeplastbox]{flushend}

\usepackage{graphicx}
\usepackage{subcaption}

\def\footnoterule{\relax%
  \kern-5pt
  \hbox to \columnwidth{\vrule width 0.4\columnwidth height 0.4pt\hfill}
  \kern4.6pt}

\newif\iffinal
\finaltrue

\begin{document}

\title{Pay Voice: Point of Sale Recognition \\ for Visually Impaired People}

\iffinal
	\author{\IEEEauthorblockN{Guilherme Folego\,$^{*}$\thanks{$^{*}$Corresponding author: \tt\href{mailto:gfolego@cpqd.com.br}{gfolego@cpqd.com.br}}\qquad
                              Filipe Costa\qquad
                              Bruno Costa\qquad
                              Alan Godoy\qquad
                              Luiz Pita}\medskip
            \IEEEauthorblockA{CPqD, Campinas, Brazil}}
\fi

\maketitle

\begin{abstract}
Millions of visually impaired people depend on relatives and friends to perform their everyday tasks.
One relevant step towards self-sufficiency 
is to provide them with means to verify the value and operation presented in payment machines. In this work, we developed and released a smartphone application, named \iffinal\emph{Pay Voice}\else\emph{XYZ}\fi, that uses image processing, optical character recognition (OCR) and voice synthesis to recognize the value and operation presented in POS and PIN pad machines, and thus informing the user with auditive and visual feedback. The proposed approach presented significant results for value and operation recognition, especially for POS, due to the higher display quality. Importantly, we achieved the key performance indicators, namely, more than $80\%$ of accuracy in a real-world scenario, and less than $5$ seconds of processing time for recognition.
\iffinal\emph{Pay Voice}\ \else\emph{XYZ}\ \fi is publicly available on Google Play and App Store for free.
\end{abstract}



\section{Introduction}
\label{sec:intro}

In the last decade, technological improvements resulted in the development of approaches to provide accessibility for people with disabilities. One important task that has been studied recently is the creation of user interfaces which are appropriate for people with visual impairment~\cite{Csapo_2015}. In Brazil, for instance, there are more than $6.5$ million people with severe permanent blindness~\cite{ibge}.

As a consequence, many assistive technologies~(AT) have been developed for helping people to identify accessible places, to communicate with individuals who have auditive problems, to describe objects for blind people, among others. One example of AT is the screen reader, a software which converts the content shown on a screen into speech. It can be useful for describing objects, informing routes, allowing blind people to read books and news, and to interact in social media. Financial operations, such as withdrawing money, paying bills and checking accounts, can also be done by blind people with the help of screen readers.

\begin{figure}[t!]
\centering

  \begin{subfigure}[b]{0.2\textwidth}
    \includegraphics[width=\textwidth]{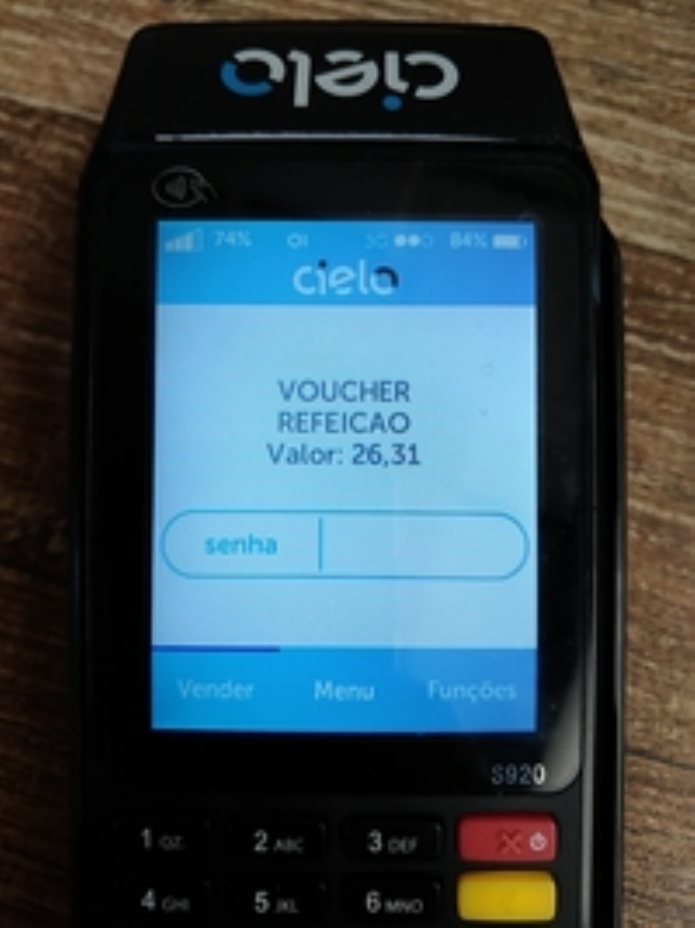}
    \caption{Point of sale.}
  \end{subfigure}
  \quad
  \begin{subfigure}[b]{0.2\textwidth}
    \includegraphics[width=\textwidth]{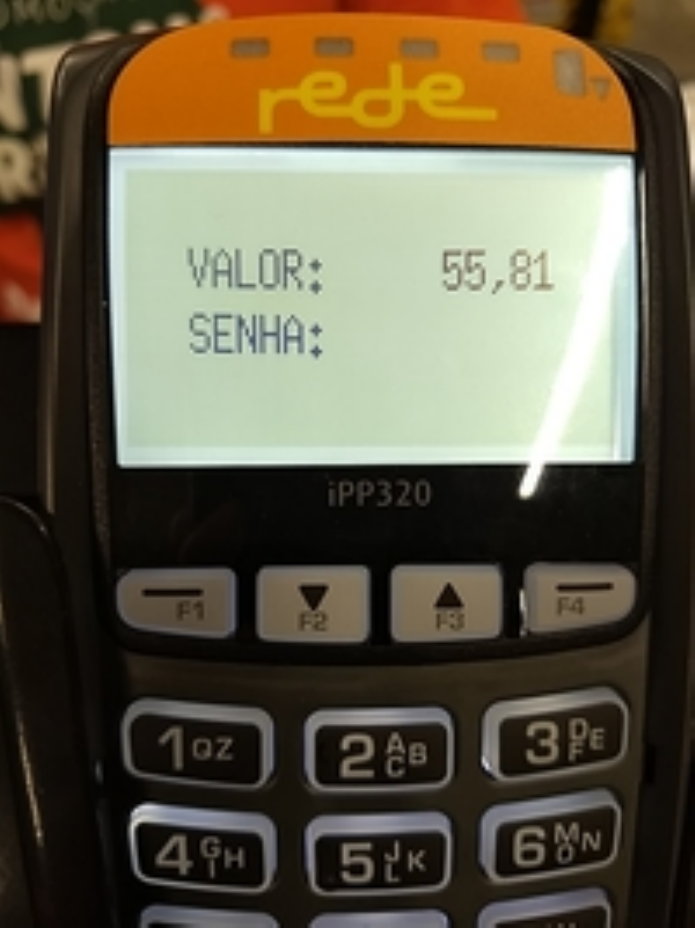}
    \caption{PIN pad.}
  \end{subfigure}
  \setlength{\belowcaptionskip}{-8pt}
  \caption{Example of payment machines.}
  \setlength{\belowcaptionskip}{0pt} 
  \label{fig:example}
\end{figure}

In this context, a challenging issue is to allow visually impaired people to check the value and the operation presented in payment machines, such as point of sale (POS) and PIN pad, illustrated in Fig.~\ref{fig:example}. While POS are stand-alone devices, PIN pads are directly connected to the merchant sales system; from a physical point of view, PIN pads are devices with simpler screens, usually backlit LCD with low resolution. One possible solution for this problem is using an embedded application in a smartphone for capturing the screen of the POS or PIN pad, recognizing the operation (\textit{e.g.}, credit, debit, voucher) and the value to be paid, converting the text to voice, and then informing the user. The main advantage of this approach is to properly work with existing payment devices in the field, which greatly reduces adoption cost. Even when these devices get upgraded in the future, this method remains viable.

However, the development of AT for reading the value presented on POS or PIN pad screens has some challenges. First, the detection of screen, value and operation to be read can be imprecise due to the environment (\textit{e.g.}, light variation, occlusion, reflection on the screen), low quality of older POS and PIN pad machines, and the positioning of the smartphone at the time of capture. Second, different brands and models of POS and PIN pad devices can have different fonts, sizes and positions for value and operation information. Finally, considering an embedded application, the processing environment can be a limiting factor, due to the timeout for completing the payment operation.

In this work, we introduce \iffinal\emph{Pay Voice}\else\emph{XYZ} \textit{[name hidden due to blind review]}\fi, a smartphone application that uses image processing, optical character recognition (OCR), and speech synthesis to recognize the value and operation presented in POS and PIN pad machines, and thus informing the user with auditive and visual feedback. The main contributions of this work are:

\begin{itemize}
\item Detection of POS and PIN pad screens in real time;
\item Detection of regions of interest in the screen;
\item Application of an OCR system to recognize the value and operation;
\item Application of speech synthesis to provide auditive feedback to the user.
\end{itemize}

\iffinal\emph{Pay Voice}\ \else\emph{XYZ}\ \fi is a software application publicly available on Google
Play\iffinal\footnote{\url{https://play.google.com/store/apps/details?id=br.org.abecs.payvoice}}\ \else\ \fi
and App Store\iffinal\footnote{\url{https://itunes.apple.com/br/app/pay-voice/id1344943724}}\fi,
which can be downloaded and used for free. Furthermore, it has received attention from the news media\iffinal~\cite{globo,band}\fi.\iffinal\else\textit{ [references hidden due to blind review]}\fi

The remaining of this paper is organized as follows. In Section~\ref{sec:related}, we present some related works, and in Section~\ref{sec:method}, we detail our proposed method. We describe our experimental setup and our dataset in Section~\ref{sec:exp}. Finally, we present and discuss our results in Section~\ref{sec:res}, and conclude the paper along with some future directions in Section~\ref{sec:conc}.


\section{Related Works}
\label{sec:related}

In order to develop this research, we evaluated a number of studies within assistive technologies and image processing areas, including OCR methods.

\subsection{Assistive technologies}

In combination with accessible user interfaces, assistive technologies can help visually impaired people in their rehabilitation and allow them to interact in the virtual world~\cite{deSouza:2017:UAA:3160504.3160583}.

According to~\cite{Csapo_2015}, the term assistive technology can be used within several approaches which require some form of assistance, and it can be divided into two main groups. The first group are approaches based on tactile solutions, which generally transform images captured by a camera into electrical or vibrotactile stimuli. It is useful for recognizing different shapes~\cite{Kaczmarek_2003}, for localization tasks~\cite{Jansson_1983}, and reading~\cite{Craig_1981}.
The second group is composed by approaches based on auditory solutions. These approaches transform the object of interest (\textit{e.g.}, video, text, image) to audio. Auditory solutions can be used, for instance, in obstacle detection~\cite{Dunai_2010}.

Another example of auditory assistive approach is text to speech~(TTS)~\cite{Piccolo:2011:DAI:2254436.2254474}, with the goal of synthesizing artificial human speech from a text. Basically, TTS approaches allow human interactions with technology without the need of visual interfaces. In general, smartphones operating systems have an integrated screen reader mechanism which uses text to speech technology to improve accessibility. For Android-based systems, Google TalkBack service allows visually impaired users to interact with their devices using audible feedback, such as spoken words and sounds. For Apple iOS-based systems there is the Voice Over screen reader. Based on gestures, it tells what the user is touching or dragging.



\begin{figure}[t!]
\centering

  \begin{subfigure}[b]{0.2\textwidth}
    \includegraphics[width=\textwidth]{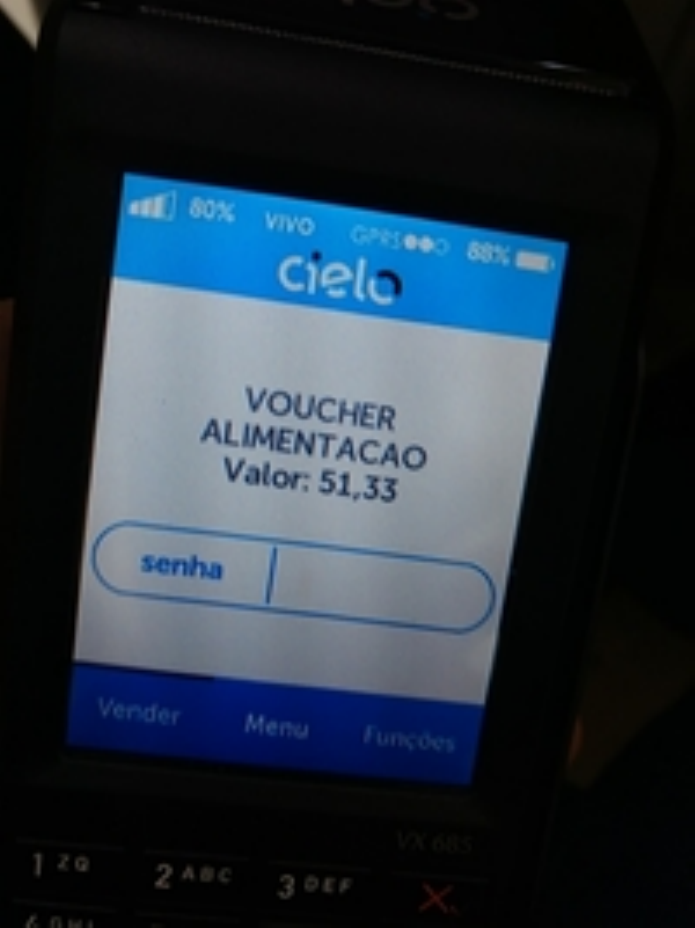}
    \caption{Input image.}
  \end{subfigure}
  \quad
  \begin{subfigure}[b]{0.2\textwidth}
    \includegraphics[width=\textwidth]{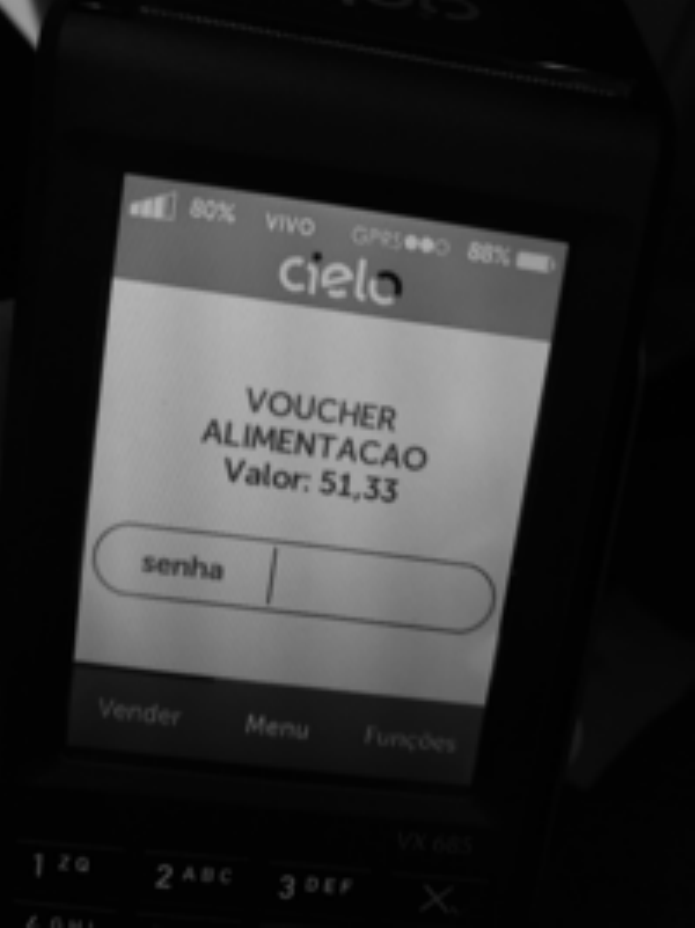}
    \caption{Converted to grayscale.}
  \end{subfigure}
  
  \vspace{.8\baselineskip}
  
  \begin{subfigure}[b]{0.2\textwidth}
    \includegraphics[width=\textwidth]{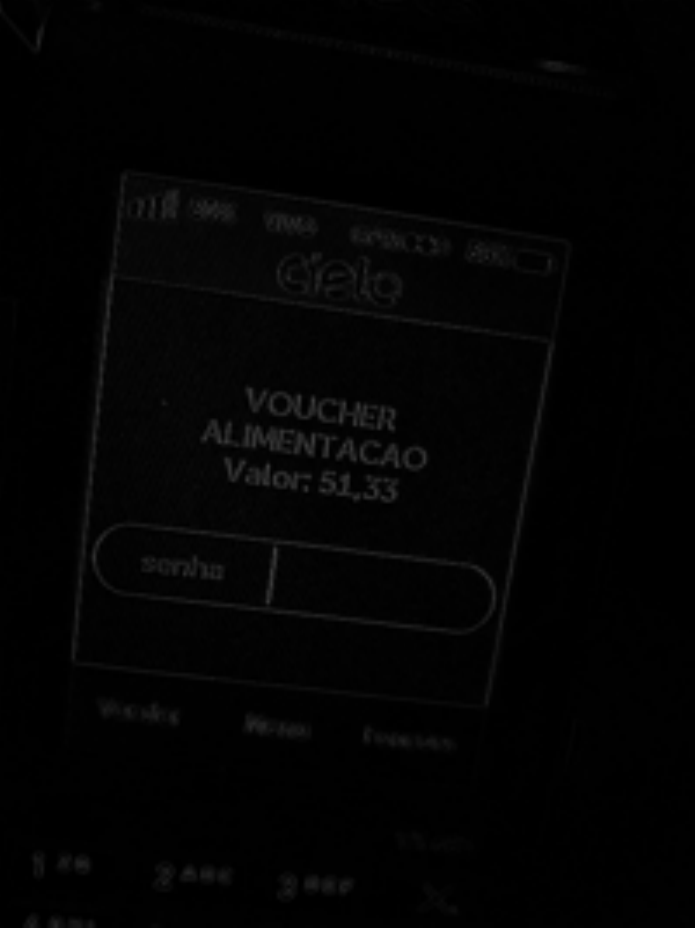}
    \caption{After Laplacian filter.}
  \end{subfigure}
  \quad
  \begin{subfigure}[b]{0.2\textwidth}
    \includegraphics[width=\textwidth]{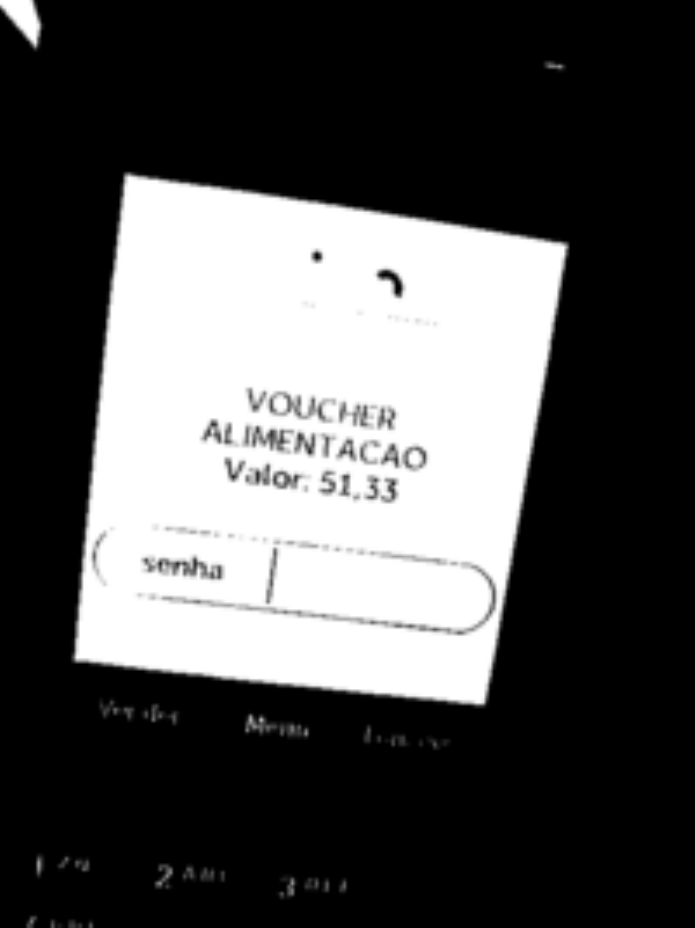}
    \caption{After Otsu threshold.}
  \end{subfigure}
  
  \vspace{.8\baselineskip}
  
  \begin{subfigure}[b]{0.2\textwidth}
    \includegraphics[width=\textwidth]{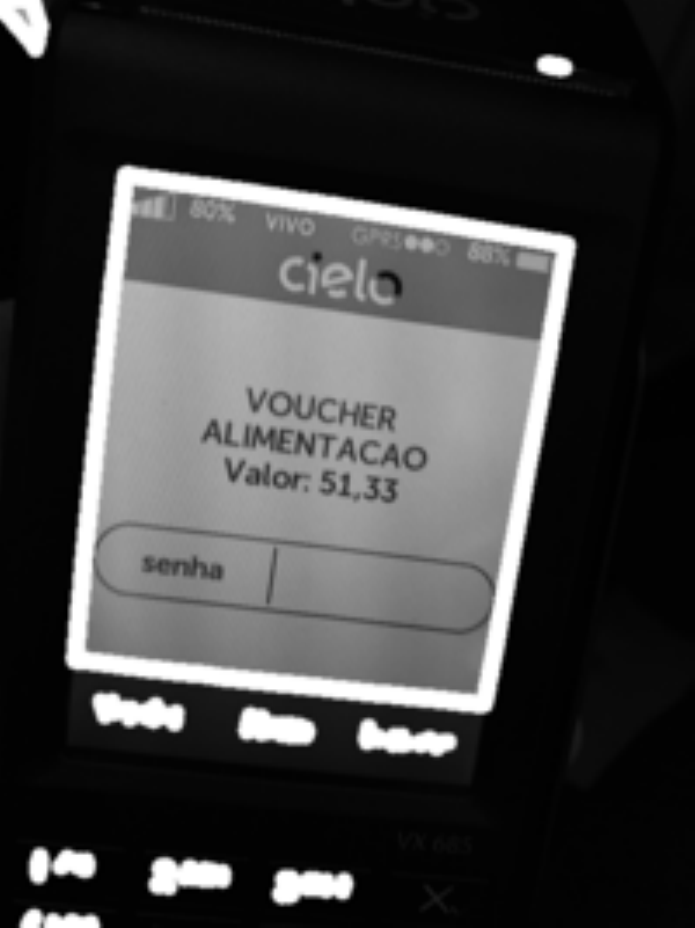}
    \caption{Found contours.}
  \end{subfigure}
  \quad
  \begin{subfigure}[b]{0.2\textwidth}
    \includegraphics[width=\textwidth]{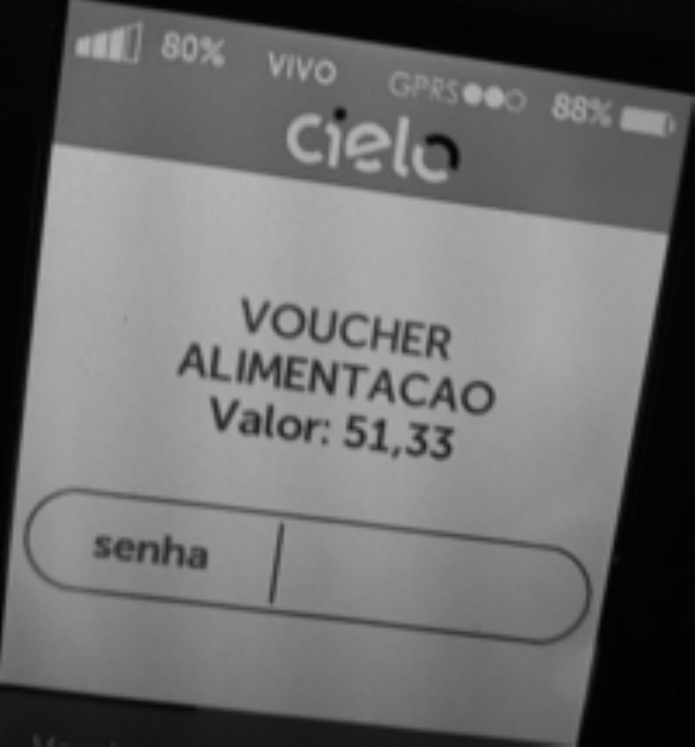}
    \caption{Detected screen.}
  \end{subfigure}
  
  \setlength{\belowcaptionskip}{-8pt}
  \caption{Screen detection steps.}
  \setlength{\belowcaptionskip}{0pt}
  \label{fig:screen}
\end{figure}


\subsection{Optical character recognition}

OCR has been studied by scientific community for several years. Although many researchers consider OCR a solved problem, the detection and recognition of text in images and videos have many challenges due to low quality or degraded data in a real-world scenario. Approaches that allow to automatically recognize characters through an optical mechanism transcribing the text have several applications in different areas, such as car plate recognition, real time translation, multimedia retrieval, and TTS methods~\cite{Ye_2015}.

One example of OCR approach is the open source OCR engine Tesseract~\cite{Smith_2007}.
Assuming the input is a binary image, the first step is to detect the outline of the components using connected component analysis. Then, the outlines are gathered together into blobs. After that, the blobs are organized into text lines, which are analyzed for fixed pitch and proportional text. The obtained lines are broken into words based on the characters spacing. Fixed pitch is chopped in character cells while proportional text is broken into words by definite spaces and fuzzy spaces.

The recognition step of Tesseract has two phases. First, the engine tries to recognize each word in turn. Each satisfactory word is passed to an adaptive classifier as training data. Once this adaptive classifier has been trained, the second phase is to run over the page trying to recognize the words which were not recognized on the first phase.


\section{Proposed Method}
\label{sec:method}

Our method consists primarily of two steps, namely, screen detection, and recognition of value and operation. The main requirement for our pipeline was to run completely embedded on smartphones, so it would not depend on Internet connection and would not suffer delays related to network latency. The screen detection step is performed in real time to provide audiovisual feedback to the user for correct positioning of the camera. After confirming we have a good image of the screen, then we perform the recognition step.

We used Tesseract as OCR engine. In order to improve recognition accuracy, we trained a specialized model to recognize texts from POS and PIN pad screens. We selected $24$ freely available fonts on the Internet to train the model.
This selection was based on the similarity to fonts present in POS and PIN pad devices. We also built a text corpus suitable for the vocabulary used in the payment context. In this section, we describe \textit{Pay Voice} version $1.5.0$.










\begin{figure}[t!]
\centering

  \begin{subfigure}[b]{0.2\textwidth}
    \includegraphics[width=\textwidth]{fig/steps05.pdf}
    \caption{Detected screen.}
  \end{subfigure}
  \quad
  \begin{subfigure}[b]{0.2\textwidth}
    \includegraphics[width=\textwidth]{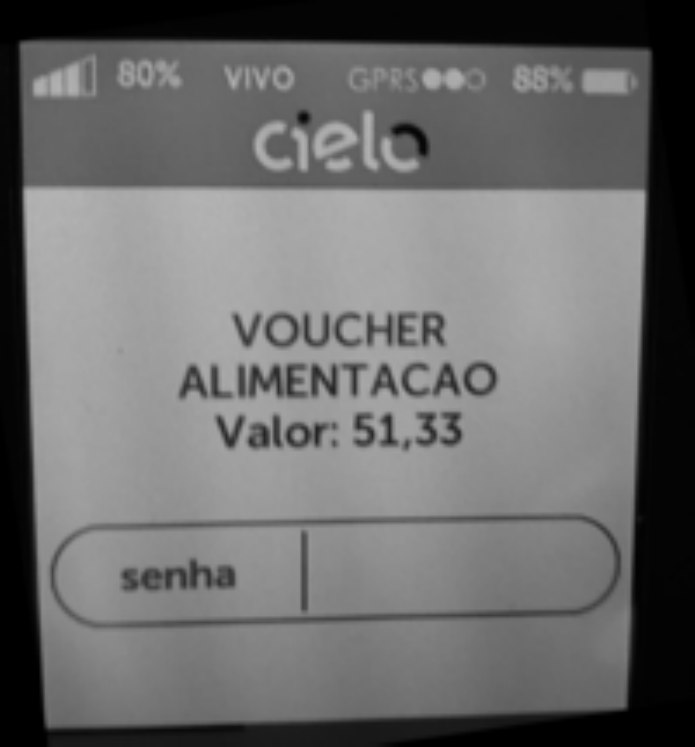}
    \caption{After rotation.}
  \end{subfigure}
  
  \vspace{.8\baselineskip}
  
  \begin{subfigure}[b]{0.2\textwidth}
    \includegraphics[width=\textwidth]{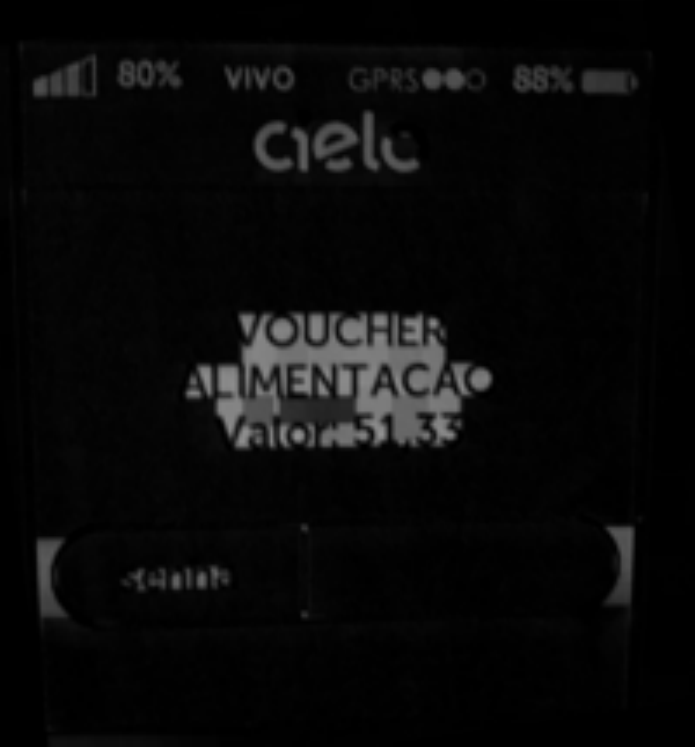}
    \caption{After top hat filter.}
  \end{subfigure}
  \quad
  \begin{subfigure}[b]{0.2\textwidth}
    \includegraphics[width=\textwidth]{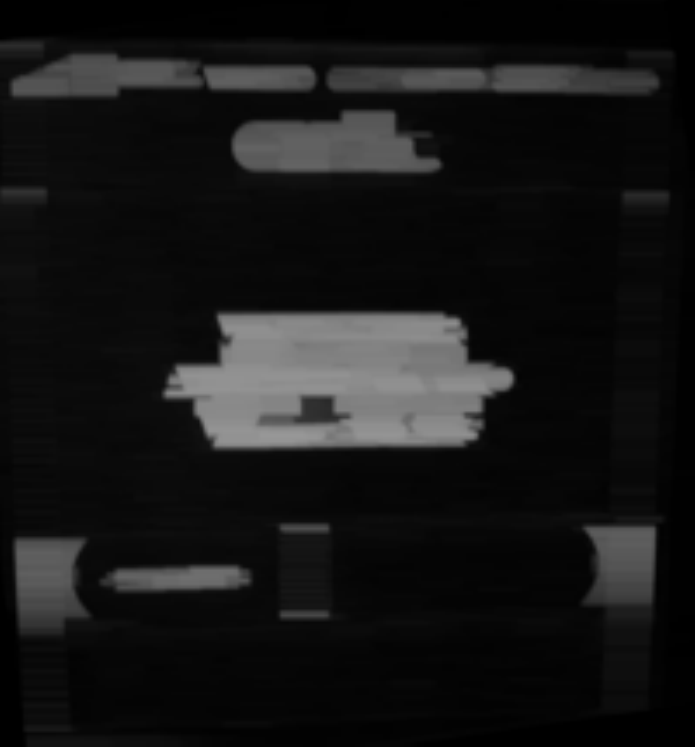}
    \caption{After dilation.}
  \end{subfigure}
  
  \vspace{.8\baselineskip}
  
  \begin{subfigure}[b]{0.2\textwidth}
    \includegraphics[width=\textwidth]{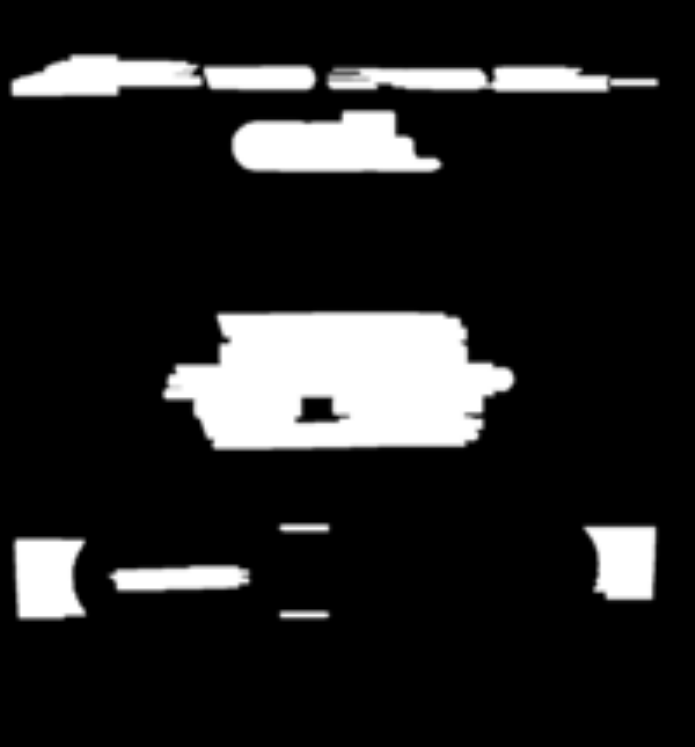}
    \caption{After Otsu threshold.}
  \end{subfigure}
  \quad
  \begin{subfigure}[b]{0.2\textwidth}
    \includegraphics[width=\textwidth]{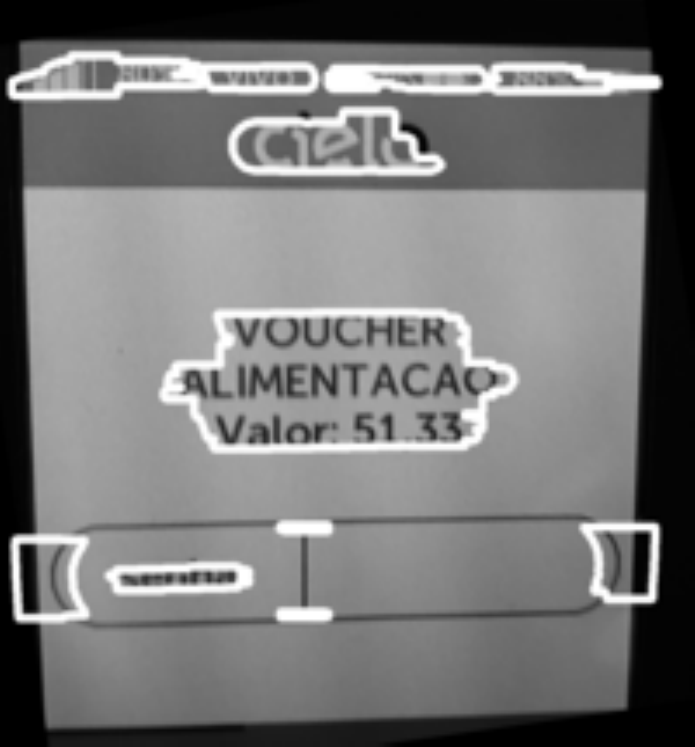}
    \caption{Found contours.}
  \end{subfigure}
  
  \setlength{\belowcaptionskip}{-8pt}
  \caption{Regions of interest detection steps.}
  \setlength{\belowcaptionskip}{0pt}
  \label{fig:regions}
\end{figure}


\subsection{Screen detection}

Since this step needs to run locally in real time, we defined some assumptions, such as the screen being the largest bright region in the image.
This demonstrated to work fairly well in our real-world experiments with a number of different machines and environments.

To ensure this step is executed as fast as necessary, we first convert the input image to grayscale, and resize it to $640$ pixels in height with nearest neighbor interpolation~\cite{Parker_1983}.
This image is used twice: to check for correct focus, and to detect the screen. The focus is checked by applying the Laplacian filter~\cite{Aubert_2006} and calculating the variance of the filtered image.
The screen is detected by applying the Otsu threshold~\cite{Vala_2013}, and searching for the contour~\cite{Suzuki_1985} with largest area.
With this contour, we compute a straight rectangle enclosing it, and project it back to the original large image, so we can use the screen region in full scale for the recognition step. We additionally compute a rectangle with minimum area around this contour, so we can estimate the screen rotation.
We present intermediate results in Fig.~\ref{fig:screen}.

In order to guarantee a good camera positioning, we provide some meaningful audiovisual feedbacks to the user.
If the area of the straight rectangle is bellow a certain threshold, then the camera is too far from the machine. Conversely, if the area is above another threshold, the camera is too close.
We also check the distance between screen rectangle borders and image edges, making sure the screen is roughly centered. It is important to note that the employed thresholds were determined empirically.
In case all these verifications are valid for five consecutive frames of the camera stream, including the focus check, then we proceed to the recognition.

\subsection{Recognition}

We start the recognition step with a median filter, to reduce the noise from the camera. Then, if the screen angle is larger than $4$ degrees (empirically determined), we rotate it around the center to the closest straight position, using an affine transformation with cubic interpolation. Since we cannot know whether the screen is vertical or horizontal, our recognition is limited to $45$ degrees of rotation. For instance, in case the screen is vertical, and it has $50$ degrees of rotation, it will end up rotated to an horizontal position, and the recognition will not be able to work properly.

After aligning the screen, we need to find the potential regions of interest. This is necessary since the OCR engine cannot handle the complete screen at once, outputting garbled text and taking a long execution time. The detection of these regions starts with resizing the screen image to half its original size with nearest neighbor interpolation, so it executes faster. We then apply a white top hat morphological operation with a large nearly square kernel, which is the difference between the input image and its opening, followed by a dilation with a large horizontal kernel, to connect nearby elements.

After, we perform an Otsu threshold and detect all contours, extracting a straight rectangle region around each of them, and we project these regions back to the original space, with an additional padding of $10$ pixels in each side. Finally, we select only the horizontal regions that have a proportional area within a predefined range, to ignore regions that are not useful and to speed the recognition. These steps are illustrated in Fig.~\ref{fig:regions}. Even though there are more general techniques for text spotting with good accuracy~\cite{10.1007/978-3-319-10593-2_34}, limitations regarding latency and hardware rendered the use of such methods impracticable in the present work.




\begin{figure}[t!]
\centering

  \begin{subfigure}[b]{0.2\textwidth}
    \includegraphics[width=\textwidth]{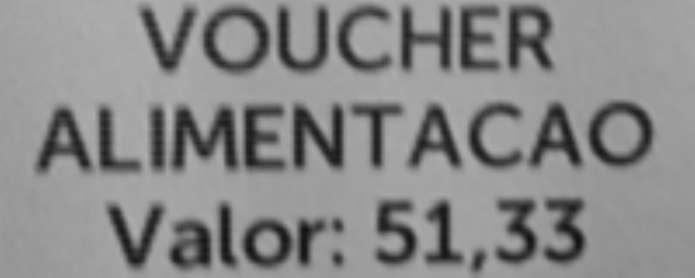}
    \caption{Detected region.}
  \end{subfigure}
  \quad
  \begin{subfigure}[b]{0.2\textwidth}
    \includegraphics[width=\textwidth]{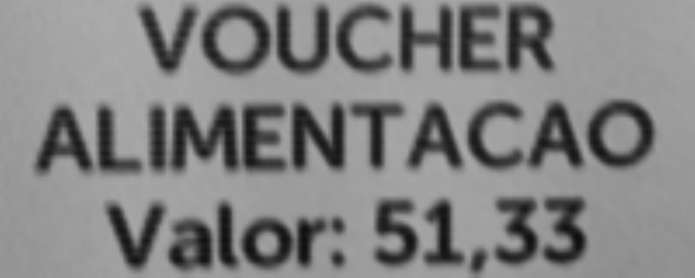}
    \caption{After dilation.}
  \end{subfigure}
  
  \vspace{.8\baselineskip}
  
  \begin{subfigure}[b]{0.2\textwidth}
    \includegraphics[width=\textwidth]{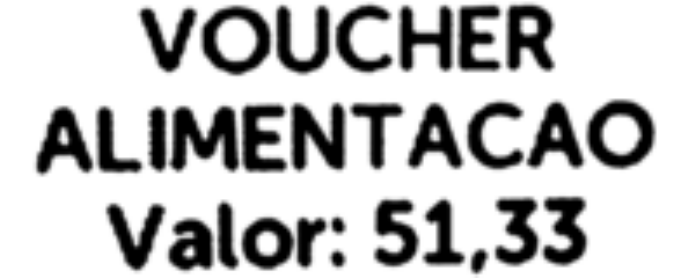}
    \caption{After Otsu threshold.}
  \end{subfigure}
  \quad
  \begin{subfigure}[b]{0.2\textwidth}
    \includegraphics[width=\textwidth]{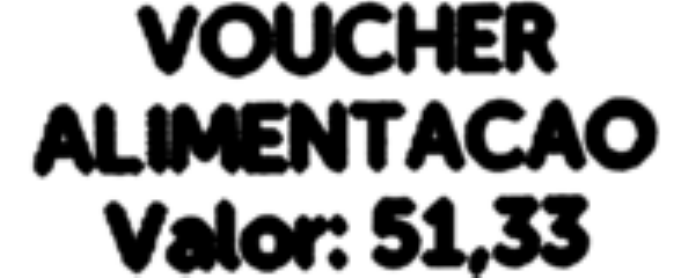}
    \caption{After dilation.}
  \end{subfigure}
  
  \setlength{\belowcaptionskip}{-8pt}
  \caption{Region filtering steps.}
  \setlength{\belowcaptionskip}{0pt}
  \label{fig:recog}
\end{figure}

\begin{figure}[t!]
\centering

  \begin{subfigure}[b]{0.2\textwidth}
    \includegraphics[width=\textwidth]{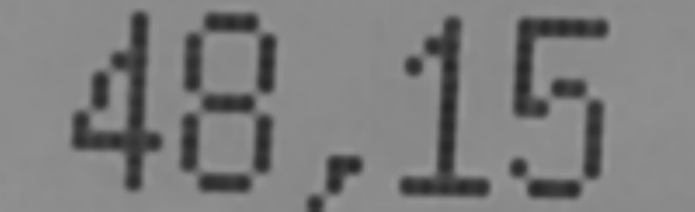}
    \caption{Detected region.}
  \end{subfigure}
  \quad
  \begin{subfigure}[b]{0.2\textwidth}
    \includegraphics[width=\textwidth]{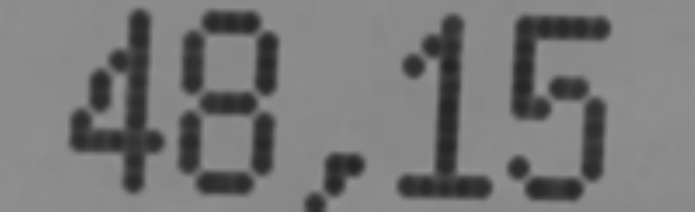}
    \caption{After dilation.}
  \end{subfigure}
  
  \vspace{.8\baselineskip}
  
  \begin{subfigure}[b]{0.2\textwidth}
    \includegraphics[width=\textwidth]{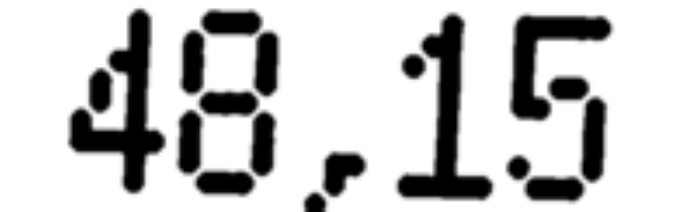}
    \caption{After Otsu threshold.}
  \end{subfigure}
  \quad
  \begin{subfigure}[b]{0.2\textwidth}
    \includegraphics[width=\textwidth]{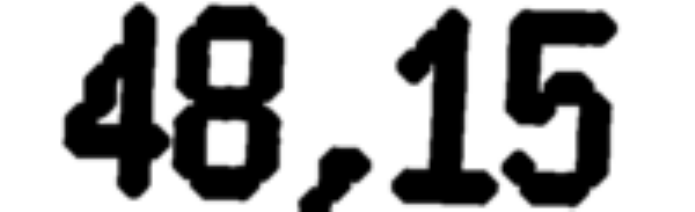}
    \caption{After dilation.}
  \end{subfigure}
  
  \setlength{\belowcaptionskip}{-8pt}
  \caption{Region filtering steps in a PIN pad machine.}
  \setlength{\belowcaptionskip}{0pt}
  \label{fig:recogpin}
\end{figure}

In the last step, for each selected region of interest, we try to recognize the value and operation.
We clean up this image using an erosion morphological operation with a small ellipse kernel, followed by an Otsu threshold, and the resulting image goes through the OCR engine. Finally, we apply another erosion, and the OCR again. These steps are illustrated in Fig.~\ref{fig:recog}. This repeated erosion and OCR process is necessary due to thin and unconnected fonts, especially in PIN pad machines, as shown in Fig.~\ref{fig:recogpin}.


After each OCR process, we try to extract the value and operation (\textit{e.g.}, credit, debit) from the recognized text. For the value, we use a regular expression that matches expressions containing integer digits, followed by a decimal separator and two integer digits. The regular expression is flexible enough to handle small variations and recognition errors (\textit{e.g.}, we accept a whitespace between the decimal separator). Confidence for value is given as the weighted mean between the score for the integer part ($0.75$) and the decimal part ($0.25$) of the value. The score for each part is computed as the average score of its characters, as given by the OCR engine. Considering all evaluated regions, we only keep the recognized value with highest calculated score.

For operation, we compute scores based on the distance between the recognized text and a set of previously known operations, selecting the one with highest score. The distance is obtained with a simplified version of Levenshtein algorithm~\cite{Lev_1965}. After that, we also compute scores based on the distance between the recognized text and a previously defined blacklist, which contains expressions we want to avoid. The blacklist contains expressions that are similar to known operations and might be confused as legitimate operations with recognition errors. For instance, the word ``\textit{digite}'' may be identified as ``\textit{débito}''.


If the score from the blacklist is higher than the score from the set of known operations, then the operation is changed to unknown. Similarly to the value, considering all evaluated regions, we only keep the operation with highest calculated score.








\begin{figure}[b!]
\centering

  \begin{subfigure}[b]{0.15\textwidth}
    \includegraphics[width=\textwidth]{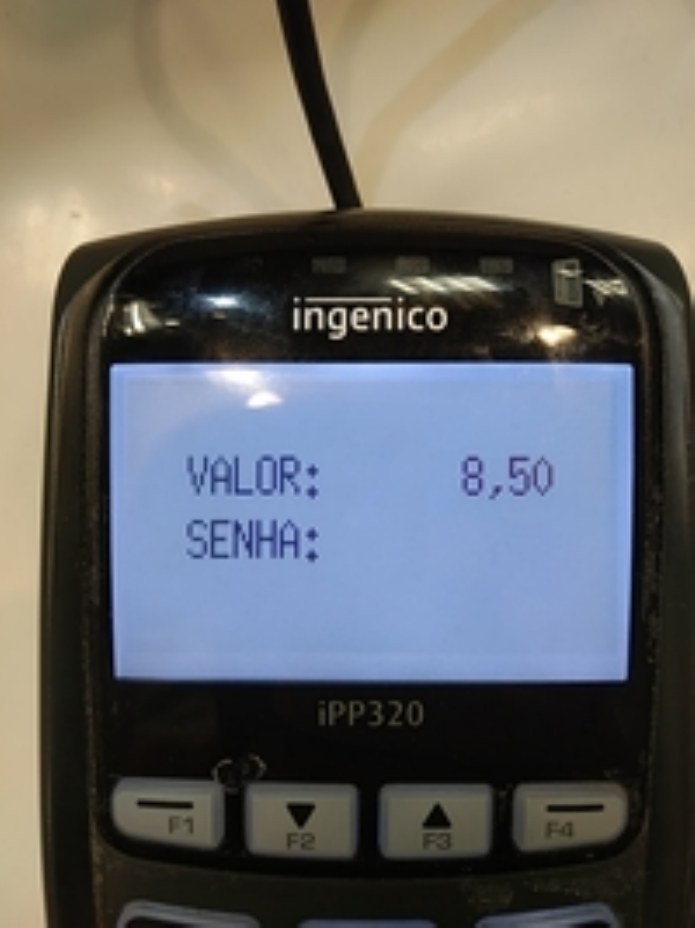}
    \caption{Simple.}
  \end{subfigure}
  \begin{subfigure}[b]{0.15\textwidth}
    \includegraphics[width=\textwidth]{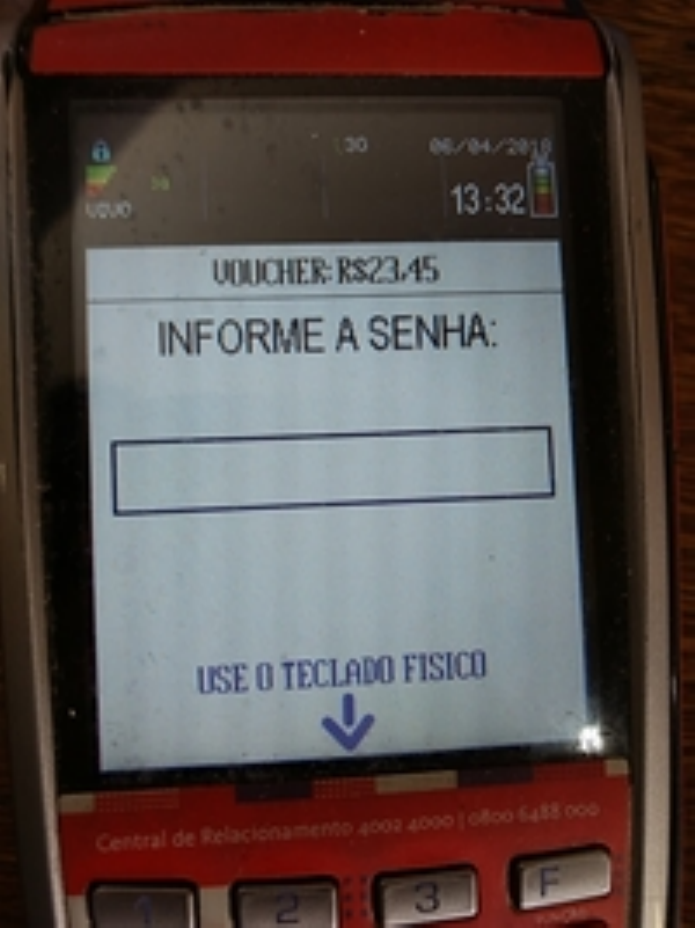}
    \caption{Difficult.}
  \end{subfigure}
  \begin{subfigure}[b]{0.15\textwidth}
    \includegraphics[width=\textwidth]{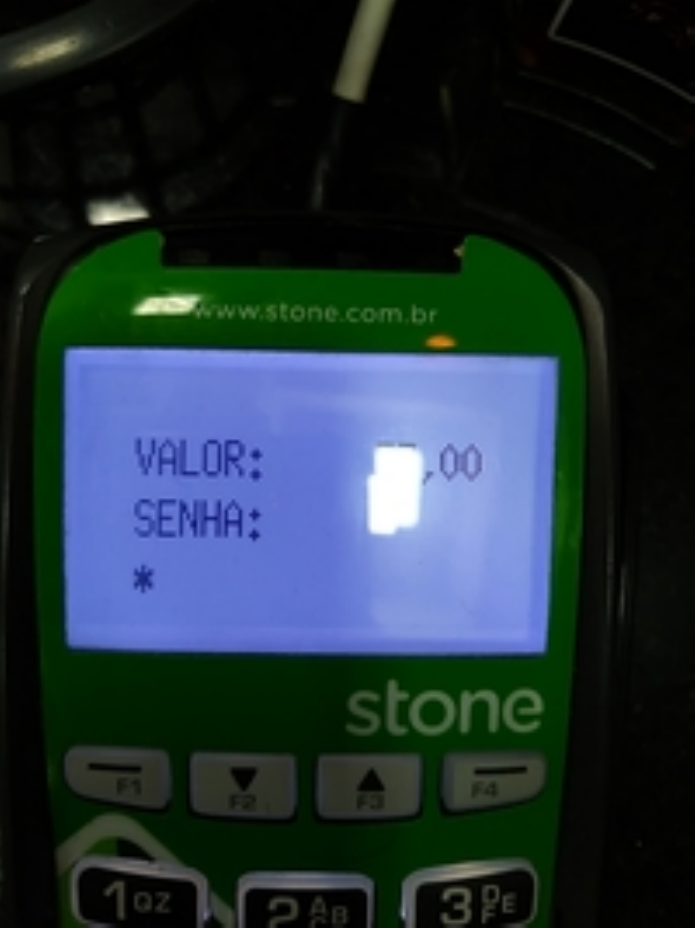}
    \caption{Impossible.}
  \end{subfigure}
  \caption{Samples from the dataset.}
  \setlength{\belowcaptionskip}{0pt} 
  \label{fig:samples}
\end{figure}



\section{Experimental Setup}
\label{sec:exp}


To validate the proposed approach, we crowdsourced the collection of a dataset in a real-world scenario, considering a variety of POS and PIN pad machines, smartphone cameras, and people performing the operation. We collected $232$ POS and $76$ PIN pad pictures, for a total of $308$ images. This collection made use of our screen detection approach, to ensure a good quality of captured images. Then, each sample was manually annotated with the respective value and operation.


\begin{figure}[t!]
\centering

  \begin{subfigure}[b]{0.15\textwidth}
    \includegraphics[width=\textwidth]{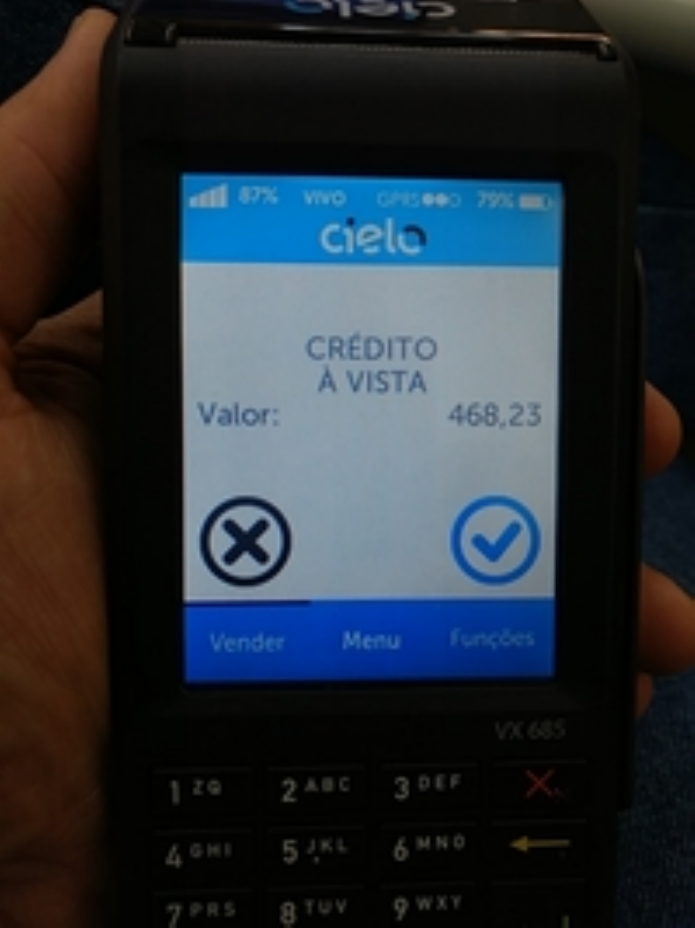}
    \caption{Original.}
  \end{subfigure}
  \begin{subfigure}[b]{0.15\textwidth}
    \includegraphics[width=\textwidth]{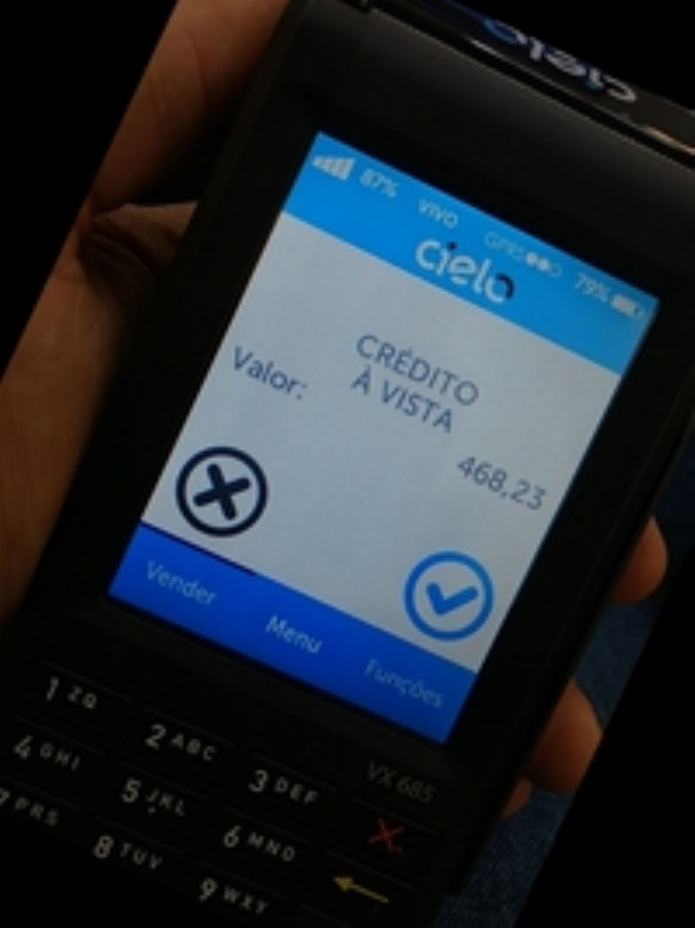}
    \caption{Cropped.}
  \end{subfigure}
  \begin{subfigure}[b]{0.15\textwidth}
    \includegraphics[width=\textwidth]{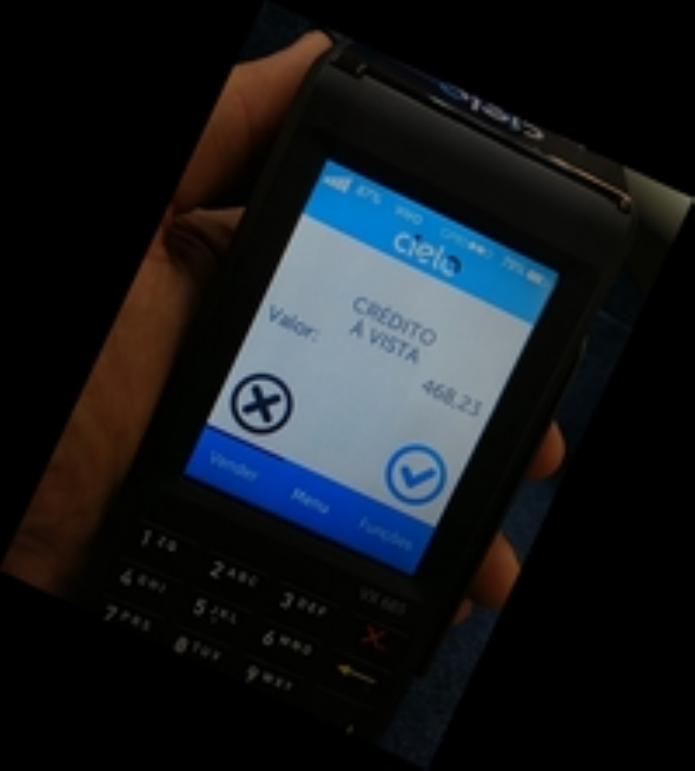}
    \caption{Rotated only.}
  \end{subfigure}
  \setlength{\belowcaptionskip}{-8pt}
  \caption{Example of simulated rotations.}
  \setlength{\belowcaptionskip}{0pt} 
  \label{fig:rot}
\end{figure}


We illustrate a simple, a difficult and an impossible sample from our dataset in Fig.~\ref{fig:samples}. Note that the simple image has few regions of interest, which makes processing faster and avoids potential mistakes. The difficult sample shows a number of regions that are larger than the value and operation, including the hours display, while the impossible image has a reflection on the value. We decided to keep every sample since they represent the real world scenario, and would better reflect on our performance metrics.
Image sizes were selected according to availability in each device, choosing the smallest size that was at least $2000$ pixels in height. Resulting sizes and respective frequencies are indicated in Table~\ref{tab:sizes}.


\begin{table}[t!]
\centering
\caption{Image sizes frequency, in pixels.}
\begin{tabular}{c*{1}{|c}}
\textbf{Size} & \textbf{Frequency} \\ \hline
$1536\times2048$	& $4$	\\
$2160\times3840$	& $1$	\\
$2448\times3264$	& $21$	\\
$2880\times3840$	& $267$	\\
$3024\times4032$	& $15$	\\
\end{tabular}
\label{tab:sizes}
\vspace{-1.5em}
\end{table}

Additionally, to evaluate our recognition with rotated images, we also generated another set of images with simulated rotations, since this was not very common in our collected data.
We selected one sample with a straight screen, and rotated it both clockwise and counterclockwise, with and without cropping. Even though we understand that our approach is limited to $45$ degrees of rotation, we generated images up to $50$ degrees in each direction, for a total of $100$ rotated images with cropping and another $100$ without cropping. The original image used is presented in Fig.~\ref{fig:rot}, along with the two generated samples with $25$ degrees of clockwise rotation.


\section{Results and discussions}
\label{sec:res}

\begin{table}[b!]
\vspace{-0.5em}
\centering
\caption{Performance of value and operation recognition, respectively.}
\begin{tabular}{c*{4}{|c}}
\textbf{Machine} & \textbf{Thr.} & \textbf{Correct} & \textbf{Incorrect} & \textbf{Unrecog.} \\ \hline
POS		& $0$	& $198\ (85.3\%)$	& $34\ (14.7\%)$	& $0\ (0.0\%)$		\\
POS		& $70$	& $198\ (85.3\%)$	& $3\ (1.3\%)$		& $31\ (13.4\%)$	\\
PIN pad	& $0$	& $59\ (77.6\%)$	& $17\ (22.4\%)$	& $0\ (0.0\%)$		\\
PIN pad	& $70$	& $58\ (76.3\%)$	& $6\ (7.9\%)$		& $12\ (15.8\%)$	\\
\end{tabular}
\label{tab:res}


\bigskip

\begin{tabular}{c*{4}{|c}}
\textbf{Machine} & \textbf{Thr.} & \textbf{Correct} & \textbf{Incorrect} & \textbf{Unrecog.} \\ \hline
POS		& $0$	& $185\ (79.7\%)$	& $47\ (20.3\%)$	& $0\ (0.0\%)$		\\
POS		& $50$	& $172\ (74.1\%)$	& $5\ (2.2\%)$		& $55\ (23.7\%)$	\\
PIN pad	& $0$	& $42\ (55.3\%)$	& $34\ (44.7\%)$	& $0\ (0.0\%)$		\\
PIN pad	& $50$	& $42\ (55.3\%)$	& $0\ (0.0\%)$		& $34\ (44.7\%)$	\\
\end{tabular}
\end{table}

Although there is a number of ways to evaluate our performance, we chose to adopt the traditional accuracy metric. In other words, we need to correctly recognize the complete value and the complete operation, considering a partially recognized value as incorrect, even if we miss only one digit.
In addition to the recognized values, our system provides a confidence score which can be used to decide whether the application will show the output from our method or will simply inform the user that it was unable to recognize the value or operation, meaning that output is shown only if scores are greater than a given threshold.

\begin{figure}[t!]
  \centering
  \includegraphics[width=0.45\textwidth]{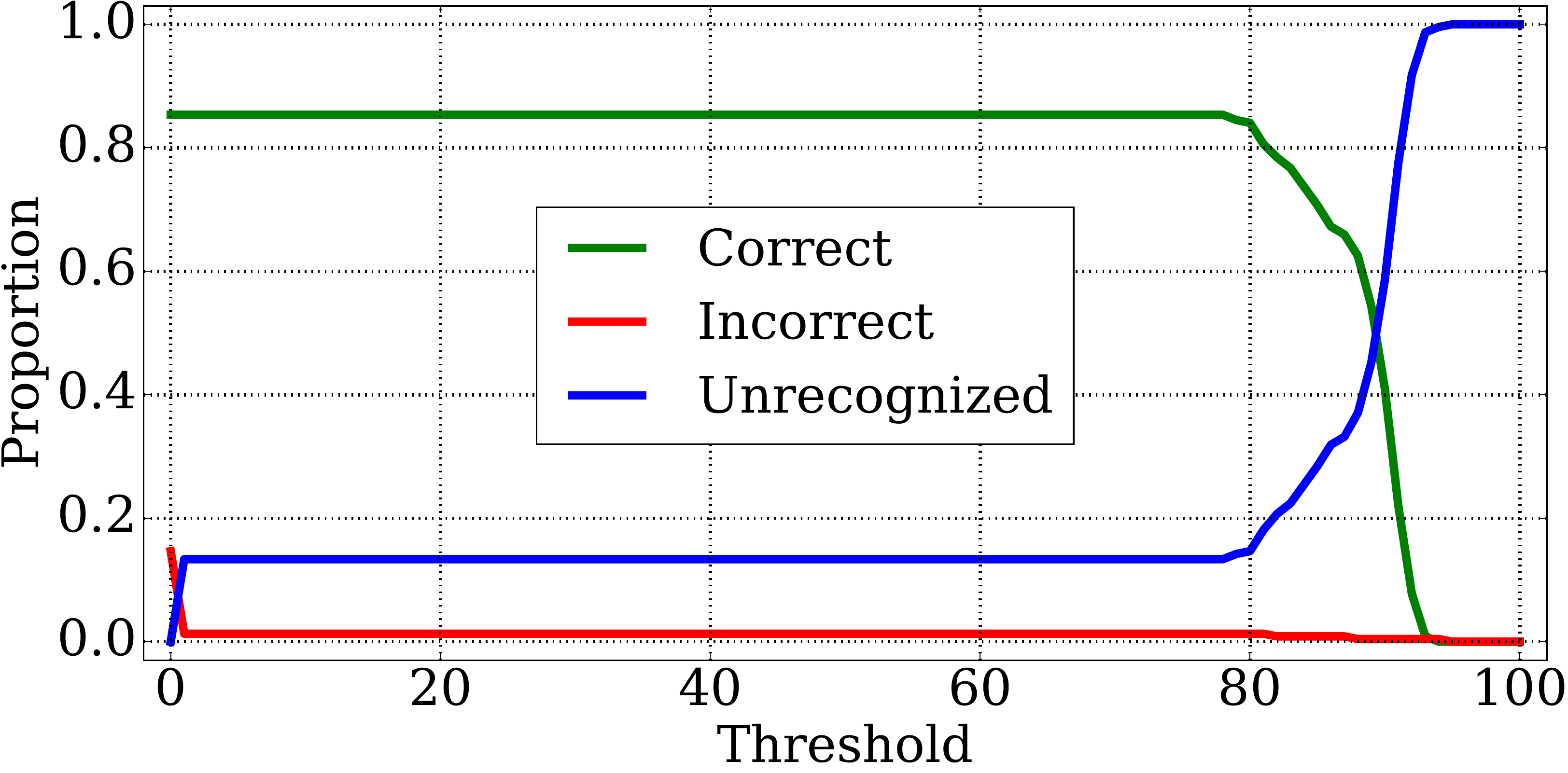}
  \setlength{\belowcaptionskip}{-8pt}
  \caption{Value recognition in POS with different thresholds.}
  \setlength{\belowcaptionskip}{0pt} 
  \label{fig:plot}
\end{figure}

Adopted thresholds, one for value and another for operation, were determined empirically.
In Fig.~\ref{fig:plot}, we plot multiple performance metrics for value recognition in POS machines, with varying thresholds, from zero to $100$. We note that most of our confidence scores are between $80$ and $90$, where there is a steep variation from correct to unrecognized values. From this curve, we opted for $70$ as a safe and sound threshold for value recognition in POS.
In Table~\ref{tab:res}, we present the performance metrics for value and operation recognition.

From Table~\ref{tab:res}, we can see that the proposed method achieved a remarkable real-world performance for POS recognition. However, PIN pad machines are somewhat harder, due to inferior illumination, higher reflection, thinner fonts and general lower quality of the screen. Using zero as threshold means we always output recognized value and operation, while higher thresholds imply eventually informing the user that this information was unrecognized. Note that some POS and almost all PIN pad machines do not display the operation, so we are simply unable to recognize it. Considering the complete dataset, in the real world ($thr = 70$), our approach correctly identified $83.1\%$ of all values, with only $2.9\%$ of incorrect recognition. Similarly for the operation ($thr = 50$), it made only $1.6\%$ of mistakes. This means that the output from our method can almost always be trusted, and reached our key performance indicator of $80\%$.

As we designed and developed this approach to work embedded in smartphones, it is important to consider how long the recognition step takes. In Table~\ref{tab:times}, we show execution times for a number of different devices. The idea was to evaluate an Android and an iOS smartphone in each tier. As such, we considered Samsung S8 and iPhone 7 as high end, LG G3 and iPhone 6 Plus as mid end, and LG X Power and iPhone 5 as low end. For comparison, we additionally included a desktop machine equipped with Intel(R) Core(TM) i7-4790 CPU @ 3.60GHz.

\begin{table}[h!]
\centering
\caption{Execution time of recognition, in seconds.}
\begin{tabular}{c*{4}{|c}}
\textbf{Device} & \textbf{Median} & \textbf{Mean} $\pm$ \textbf{StdDev} & \textbf{Min} & \textbf{Max} \\ \hline
Desktop		  & $0.378$ & $0.431 \pm 0.227$ & $0.121$ & $1.574$  \\
iPhone 7	  & $0.792$ & $0.883 \pm 0.373$ & $0.280$ & $2.995$  \\
Samsung S8	  & $1.124$ & $1.249 \pm 0.546$ & $0.386$ & $3.940$  \\
iPhone 6 Plus & $1.636$ & $1.811 \pm 0.763$ & $0.571$ & $6.446$  \\
LG G3		  & $2.548$ & $2.899 \pm 1.654$ & $0.771$ & $10.975$ \\
iPhone 5	  & $2.718$ & $2.988 \pm 1.262$ & $0.946$ & $9.799$  \\
LG X Power	  & $3.735$ & $4.212 \pm 2.045$ & $1.401$ & $16.537$ \\
\end{tabular}
\label{tab:times}
\end{table}

From Table~\ref{tab:times}, we can see that, even for the slowest device, recognition can be expected to complete in less than $5$ seconds, which completely satisfies usability requirements from our client and from our users, considering their usage experience. Even though the maximum time was around $16$ seconds, it is still acceptable, since the timeout of POS and PIN pad machines are usually $30$ seconds, and smartphone hardware tend to improve over time. As a complementary note, this maximum execution time happened due to OCR delay in unimportant regions caused by image noise.

Regarding the generated dataset with artificial rotations, our method correctly classified $88\%$ of the images with rotation only, and $89\%$ of the images with both rotation and cropping. These metrics are well within our expectations and demonstrate that our method can properly deal with rotated images, which improves the user experience.


\section{Conclusions}
\label{sec:conc}

More than 6.5 million people in Brazil show some type of blindness, relying on relatives and friends to perform ordinary tasks, as shopping and managing their personal finances. An important task to improve self-sufficiency of visually impaired people is to provide them with means to interact with payment machines. In this work, we developed and released a mobile application, named \iffinal\emph{Pay Voice}\else\emph{XYZ}\fi, for recognition of value and operation in POS and PIN pad machines in the real world, focused on helping visually impaired people. The proposed approach presented significant results for value and operation recognition, especially for POS, due to the higher display quality. Importantly, we achieved the expected results from our client, namely, more than $80\%$ of accuracy in a real world setting, and less than $5$ seconds of processing time for recognition.
We understand that this work is simply one step towards promoting integration and accessibility to visually impaired people, making them more independent.

Our intended next steps focus mainly in improving recognition performance and speed. In particular, there is a need to increase recognition accuracy for PIN pad machines. We expect that optimizing the OCR system with additional fonts, closely related to the ones we are currently missing, could bring this improvement.
Additionally, there is a number of approaches to reduce execution time. For instance, if we detect that the region of interest contains white text on black background, it is most certainly a POS, and we can safely avoid the repeated dilation and OCR steps. In order to improve eventual OCR delays, we could remove border leftovers and small artifacts from the regions of interest before proceeding with the OCR, and eventually adding a timeout to the OCR processing. Increasing our blacklist could also  potentially avoid mistakes and unnecessary steps. Finally, in principle, regions of interest processing could be done in parallel. $\blacksquare$


\begin{quote}
For most people technology makes things easier. For people with disabilities, however, technology makes things possible. \hfill (Mary Pat Radabaugh)
\end{quote}



\iffinal
\section*{Acknowledgment}
We thank the support from Claudinei Martins, Rodrigo Morbach, Fernando Marino, and Fabiani de Souza. The resulting \emph{Pay Voice} application would not be possible without the effort from all the people involved in this project.
We also appreciate the financial support from Abecs (Associação Brasileira das Empresas de Cartões de Crédito e Serviços).
\fi

\bibliographystyle{IEEEtran}
\bibliography{paper}

\end{document}